\documentclass[11pt]{article}
\usepackage{coling2020}
\usepackage{times}
\usepackage{url}
\usepackage{latexsym}

\usepackage{color}
\usepackage{comment}
\usepackage{booktabs}
\usepackage{graphicx}
\usepackage{multirow}
\usepackage[whole]{bxcjkjatype} %
\usepackage{tikz}
\usepackage{xcolor}
\usepackage{colortbl}
\usepackage[multiple]{footmisc}

\newcounter{magicrownumbers}
\newcommand\rownumber{\small{\stepcounter{magicrownumbers}\arabic{magicrownumbers} :}}
\newcommand{\fc}[1]{\color{blue} #1}

\colingfinalcopy %

\title{PheMT: A Phenomenon-wise Dataset for Machine Translation Robustness on User-Generated Contents}

\author{Ryo Fujii$^{1}$, Masato Mita$^{2,1}$, Kaori Abe$^{1}$, Kazuaki Hanawa$^{2,1}$, Makoto Morishita$^{3,1}$, \\
\bf{Jun Suzuki$^{1,2}$, Kentaro Inui$^{1,2}$} \\
$^{1}$Tohoku University, $^{2}$RIKEN, $^{3}$NTT Communication Science Laboratories \\
  
{\tt \{r-fujii, abe-k, jun.suzuki\}@ecei.tohoku.ac.jp,} \\
{\tt \{masato.mita, kazuaki.hanawa\}@riken.jp,} \\
{\tt makoto.morishita.gr@hco.ntt.co.jp,} \\
{\tt inui@ecei.tohoku.ac.jp} \\}

\date{}

\begin{document}
\maketitle
\begin{abstract}
  Neural Machine Translation (NMT) has shown drastic improvement in its quality when translating clean input, such as text from the news domain. However, existing studies suggest that NMT still struggles with certain kinds of input with considerable noise, such as User-Generated Contents (UGC) on the Internet. To make better use of NMT for cross-cultural communication, one of the most promising directions is to develop a model that correctly handles these expressions.
Though its importance has been recognized, it is still not clear as to what creates the great gap in performance between the translation of clean input and that of UGC. To answer the question, we present a new dataset, \textbf{PheMT}, for evaluating the robustness of MT systems against specific linguistic phenomena in Japanese-English translation. Our experiments with the created dataset revealed that not only our in-house models but even widely used off-the-shelf systems are greatly disturbed by the presence of certain phenomena.

\end{abstract}

\section{Introduction}
\label{intro}

\blfootnote{
    \hspace{-0.65cm}  %
    This work is licensed under a Creative Commons 
    Attribution 4.0 International License.
    License details:
    \url{http://creativecommons.org/licenses/by/4.0/}.
}

The advancement of Neural Machine Translation (NMT) has brought great improvement in translation quality when translating clean input, such as text from the news domain~\cite{luong:2015:emnlp,vaswani:2017:nips}, and it was recently claimed that NMT has even achieved human parity in certain language pairs~\cite{hassan:2018,barrault:2019:wmt}.
Despite its remarkable advancements, the applicability of NMT over User-Generated Contents (UGC), such as social media text, still remains limited~\cite{michel:2018:emnlp,berard:2019:wngt}.
Since UGC are prevailing in our real-life communication, it is undoubtedly one of the challenges we need to overcome to make MT systems invaluable for promoting cross-cultural communication.

\begin{figure}[t]
\begin{tabular}{cc}
\begin{minipage}[t]{0.33\hsize}
\centering
\includegraphics[keepaspectratio, scale=0.23]{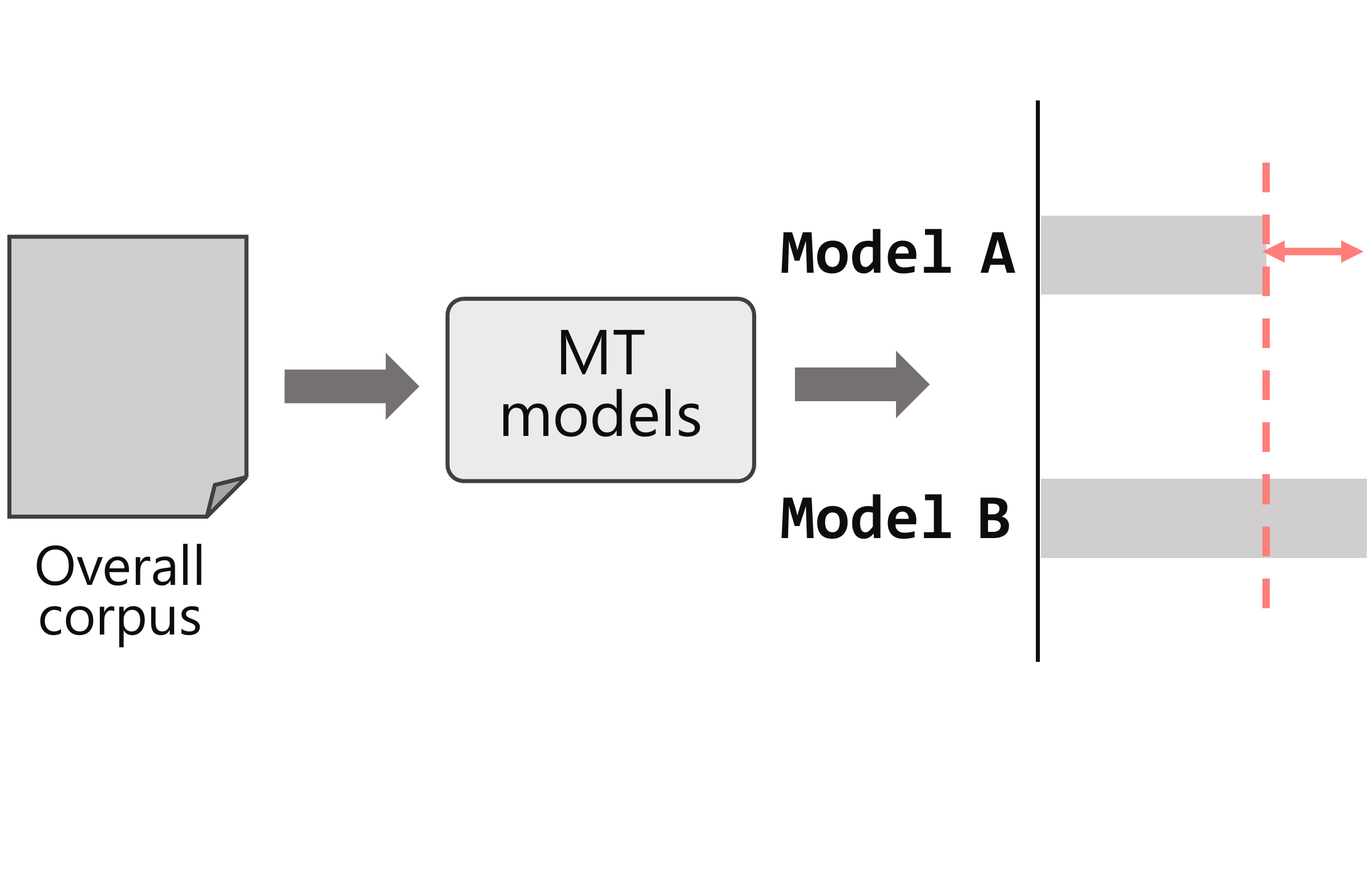}

\small{(a) Traditional MT evaluation}
\end{minipage} &
\begin{minipage}[t]{0.67\hsize}
\centering
\includegraphics[keepaspectratio, scale=0.23]{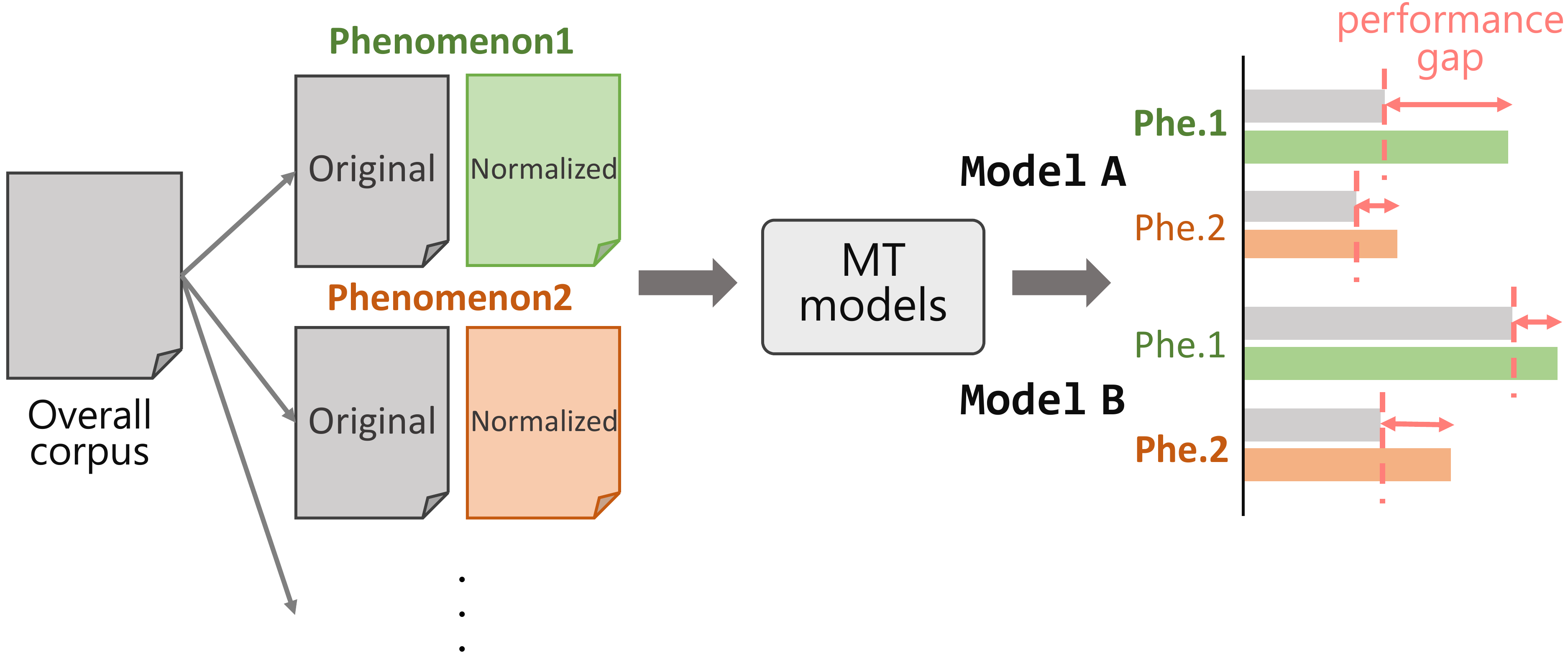}

\small{(b) Phenomenon-wise evaluation (Ours)}
\end{minipage} 
\end{tabular}
\caption{Overview of traditional MT evaluation and our proposal: phenomenon-wise evaluation.}
\vskip -2mm
\label{fig:overview}
\end{figure}

Recently, with the increasing interest in handling UGC, a shared task was organized to measure how well MT systems adapt to those texts~\cite{li:2019:wmt}. 
However, the way in which they evaluate systems is just giving an overall score to a dataset, which is the same as traditional MT evaluation (Figure~\ref{fig:overview}a). 
The overall score does not provide precise information for understanding what leads to the huge performance gap between the translation of clean input and that of UGC. 
To find a clue for improving the performance of MT systems on UGC, we need a solid basis for more detailed error analysis. 

As a first step towards a more refined evaluation of MT systems on UGC, we present a new dataset, \textbf{PheMT}: \textbf{Phe}nomenon-wise Dataset for \textbf{M}achine \textbf{T}ranslation Robustness, designed for phenomenon-wise evaluation in Japanese-English translation (Figure~\ref{fig:overview}b).
More specifically, we create a set of datasets, each of which provides a focused evaluation on one of four linguistic phenomena commonly seen on UGC, i.e., \textit{Proper Noun}, \textit{Abbreviated Noun}, \textit{Colloquial Expression} and \textit{Variant}.
By focusing locally on a specific part of a sentence presenting one of the above phenomena, we directly measure the ability of MT systems to handle the phenomenon with the help of translation accuracy.
Moreover, based on the idea of contrastive datasets, we normalize targeted expressions to its canonical form in the dictionary.
We feed both original and normalized versions of a source sentence to obtain the difference of arbitrary metrics as our robustness measure.
Using our dataset, we analyze the strengths and weaknesses of current NMT systems from the point of available training data size and the way of tokenization.
We reveal that some of the phenomena are severely problematic even to widely used, strong off-the-shelf systems.

We made our dataset publicly available for further development in MT systems.
We hope our dataset will provide promising directions to future MT systems and move the community one step forward with an additional axis for evaluation.

The contributions of this paper are:

\begin{enumerate}
    \setlength{\itemsep}{0cm}
    \item{We proposed a novel dataset designed for phenomenon-wise evaluation in Japanese-English translation as a protocol for detailed error analysis.}
    \item{We revealed with our dataset that some of the phenomena commonly seen on UGC greatly degrade the performance of current NMT systems, including widely used off-the-shelf systems.}
\end{enumerate}

\section{Related Work}

Michel and Neubig~\shortcite{michel:2018:emnlp} created the MTNT dataset with the increasing interest in creating noise-robust MT systems.
They collected comments from the social discussion website, Reddit\footnote{\url{https://www.reddit.com}}, and added translations by professional translators.
They also provided statistics of the dataset, showing that the source side of the dataset is much noisier than previous benchmarks for MT systems.
Their results with the baseline systems demonstrated the difficulty of properly translating UGC.
The dataset was also used as in-domain data for the first shared task on machine translation robustness held at WMT 2019.\footnote{\url{http://www.statmt.org/wmt19/robustness.html}}

However, the dataset is still miscellaneous in the degree of politeness, domain of the conversations, and even in the quality of translations.
Though it is still a question whether we actually need to develop any UGC-specific techniques or not, we do not even know with such a many-sided dataset that how much the improvement in some metrics, such as BLEU score~\cite{papineni:2002:acl}, actually contributes to improve robustness on various noise.
In fact, Berald et al.~\shortcite{berard:2019:wmt}, the winning team in the shared task, reported that none of the techniques specifically designed for UGC was more effective in improving BLEU score than corpus filtering.
Though there is no doubt that corpus filtering is one of the essential techniques for data-driven MT systems~\cite{koehn:2018:wmt,junczys-dowmunt:2018:wmt}, this is rather aimed at removing inappropriate \textit{sentence pairs} generated during the process of creating corpora, not at handling noisy input.
The way of current evaluation definitely prevents us from developing truly robust systems, and motivated us to create a new dataset for focused evaluation.

A range of studies have aimed to elucidate the cause of mistranslations from the viewpoint of linguistic phenomena, such as typographical errors~\cite{heigold:2018:amta,yonatan:2018:iclr,karpukhin:2019:wnut,niu:2020:acl}, grammaticality~\cite{sennrich:2017:eacl}, presence of named entities~\cite{ugawa:2018:coling}, and identification of anaphoric pronouns~\cite{bawden:2018:naacl,muller:2018:wmt}.

One of the pioneering works to analyze the behavior of NMT is the challenge set approach proposed by Isabelle et al.~\shortcite{isabelle:2017:emnlp}.
They defined various subcategories of structural differences between English and French to evaluate how well models can handle them in detail.
Though the approach has the potential of accelerating our understanding of MT systems, there lies a problem that their way of evaluation requires human evaluators with highly advanced knowledge on linguistics.

In response to the problem, Sennrich~\shortcite{sennrich:2017:eacl} proposed the contrastive dataset approach to automatically evaluate the grammaticality of a model in a comparative manner.
The author added an error-introduced contrastive version of reference to each source sentence by minimally modifying gold reference translations.
They defined the accuracy of a model as the number of times the model assigned a higher conditional probability to the original reference.
The approach was later followed by Bawden et al.~\shortcite{bawden:2018:naacl} to evaluate models' ability to exploit preceding contexts.
However, as the authors pointed out, the evaluation does not guarantee that the most probable translation by the model is free from errors even if the model ranked two references correctly.

Similar but different way of contrastive evaluation is performed on a clean input and its noisy counterpart.
Heigold et al.~\shortcite{heigold:2018:amta} introduced rule-based character replacement noise to imitate misspellings found in a variety of real-world applications.
Following work by Karpukhin et al.~\shortcite{karpukhin:2019:wnut} and Belinkov and Bisk~\shortcite{yonatan:2018:iclr} extended its scope to natural noise by using edit histories from online websites.
However, instead of giving translations to raw noisy sentences, they relied on a noisy version of input artificially created from the clean text.
Anastasopoulos et al.~\shortcite{anastasopoulos:2019:naacl} is similar to our work in that they explored the effect of errors naturally created by humans.
They focused on the effect of grammatical errors against NMT by adding translations to the JFLEG corpus~\cite{napoles:2017:eacl}, one of the common benchmarks for grammatical error correction.
Their results demonstrated that even a very small perturbation could significantly drop the performance of MT systems while exposure to similar noise during training time alleviates the problem.

However, these aspects are only a small subset of possible reasons to explain why current models are still not good at handling UGC.
To the best of our knowledge, there is no previous work aimed at investigating the effect of UGC-specific challenges in a fine-grained manner.
Also, behavioral analysis of NMT in dissimilar language pairs such as Japanese-English has not been studied extensively.
We expect a brand-new solution in this challenging language pair to be developed in the future with our dataset.

\begin{figure}[t]
    \centering
    \includegraphics[width=0.95\linewidth]{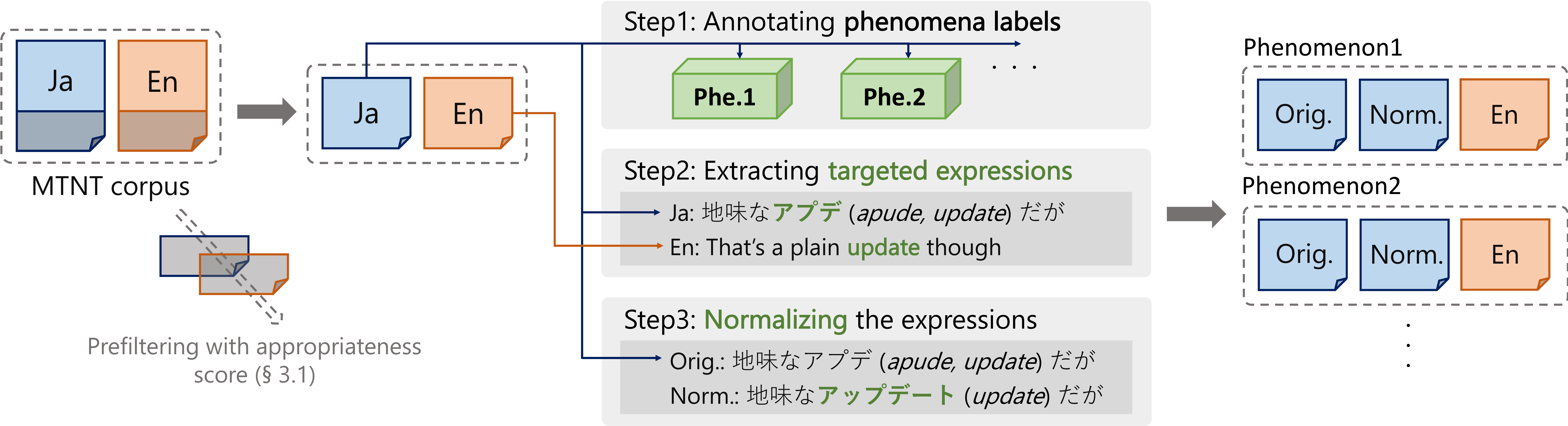}
    \caption{Entire flow of our phenomenon-wise dataset creation.}
    \vskip -2mm
    \label{fig:dataset_creation}
\end{figure}

\section{Creating Phenomenon-wise Dataset}\label{sec:dataset}

\subsection{Data selection for quality assurance}\label{subsec:approp_anno}
\begin{figure}[t]
    \centering
    \includegraphics[width=0.9\linewidth]{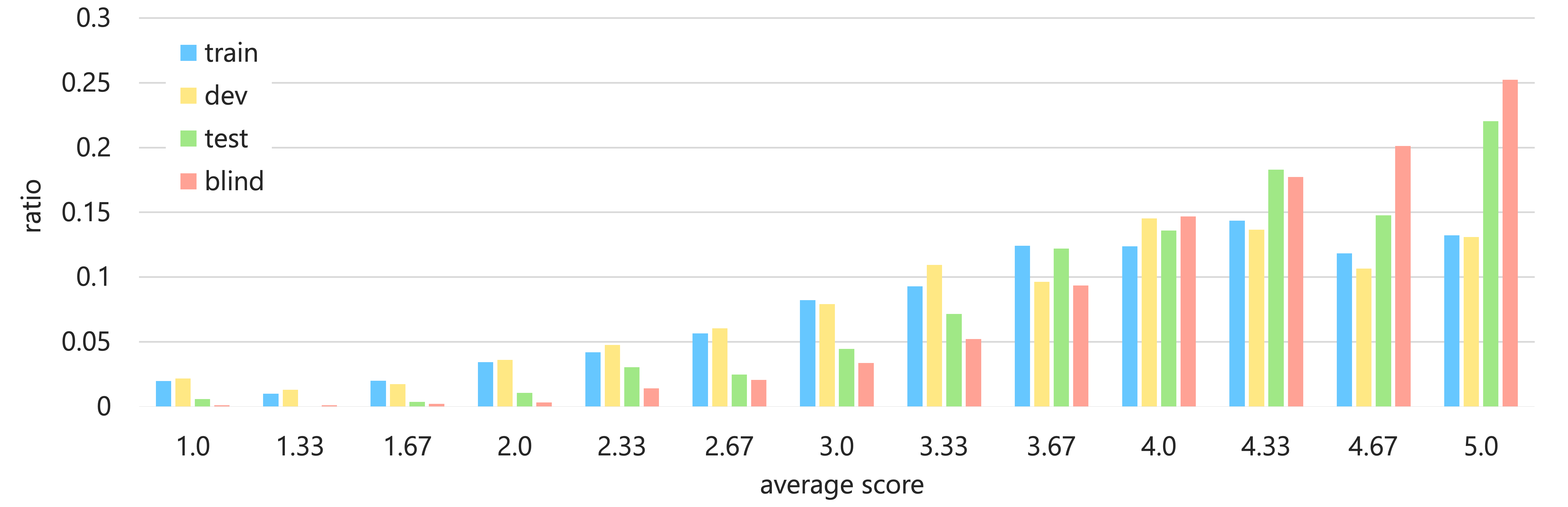}
    \caption{Distribution of appropriateness scores for the MTNT dataset. Human evaluators answered the question on the basis of a 5-point scale: 1 (very poor) -- 5 (excellent).}
    \vskip -2mm
    \label{fig:appropriateness}
\end{figure}

The entire flow of our dataset creation is described in Figure~\ref{fig:dataset_creation}.
As the methodology to create brand-new, high-quality parallel data for UGC is not trivial, we started with the existing dataset for machine translation robustness, the MTNT dataset~\cite{michel:2018:emnlp}.
The number of sentences originally created for evaluation was not enough to be further classified into several categories, so we have decided to utilize the train and development data as well to scale out our dataset.
However, such data might not be of sufficient quality to be adopted as evaluation data.
To confirm how much low-quality data it actually contains, we manually assessed the \textit{appropriateness} of source-target sentence pairs in the dataset as a preliminary experiment.{\footnote{See Appendix~\ref{appendix:appropriateness_exp} for detailed experimental settings.}}
Figure \ref{fig:appropriateness} shows the distribution of annotated scores for the MTNT dataset.
We filtered out sentences by the threshold of 4.0 to assure the quality of our phenomenon-wise dataset.

\subsection{Annotation of linguistic phenomena}
\label{subsec:pheno_anno}
\paragraph{(i) Definition of phenomena labels}
To define the labels, we first investigated what kind of UGC-specific phenomena cause significant errors in other NLP applications.
Sasano et al.~\shortcite{sasano:2013:ijcnlp} and Saito et al.~\shortcite{saito:2014:coling} focused on the presence of unnormalized orthographic variations in Japanese morphological analysis.
They introduced some handcrafted derivation rules, such as inserting prolonged sounds and substitutions to lowercased characters, to simulate alternate forms typically seen on the Internet.
Ikeda et al.~\shortcite{ikeda:2016:wnut} also applied similar rules to create synthetic data for text normalization task and demonstrated its effectiveness in improving the robustness of neural-based models.
However, the impact of those expressions has yet to be explored in a variety of cross-lingual tasks including machine translation.
Thus in this paper, we defined two types of linguistic phenomena, namely \textit{Colloquial Expression} and \textit{Variant}, by following their derivation rules.

Additionally, we defined \textit{Proper Noun} and \textit{Abbreviated Noun}, two phenomena commonly seen across various domains including UGC.
To estimate how many of the sentences in UGC actually contains these phenomena, we randomly selected 500 sentences from the training data of the MTNT dataset and annotated them in our preliminary experiment.
The result showed that more than 40\% of the sentences included one or more proper nouns, and more than 10\% of the sentences had abbreviated nouns.
Also, the effect of named entities over machine translation is receiving more and more attention in the context of transliteration~\cite{shao:2016:news,rosca:2016}.

To summarize, we targeted four phenomena as described below in our phenomenon-wise dataset (see Table~\ref{tab:label_example} for examples) ;

\begin{itemize}
    \setlength{\itemsep}{0.1cm}
    \item {\textit{Proper Noun} ; the name of a person, company, country and others, something that is unique.}
    \item {\textit{Abbreviated Noun} ; nouns made by abbreviating its canonical form, including acronyms.}
    \item {\textit{Colloquial Expression} ; words deviated from its canonical form by inserting/dropping/replacing vowels, consonants, prolonged sounds (``ー''), or geminate consonants (``っ'').}
    \item {\textit{Variant} ; words deviated from its canonical form by lowercasing characters, or by using unusual \textit{hiragana}, \textit{katakana} notation.}
\end{itemize}

\begin{table*}[t]
\centering
\scalebox{0.9}{
\begin{tabular}{@{}ll@{}}
\toprule
\bf Annotation label &\bf Examples \\ \midrule 
\textit{Proper Noun} & \small{安倍首相 (\textit{abeshush\=o}, Prime Minister Abe), 平昌 (\textit{Pyongchang})} \\
\rowcolor{gray!7}
\textit{Abbreviated Noun} & \small{アプデ (\textit{apude}, update), WHO (World Health Organization)} \\
\textit{Colloquial Expression} & \small{ねむーーい (\textit{nem\=ui}, sleepy), かなちい (\textit{kanachii}, sad)} \\
\rowcolor{gray!7}
\textit{Variant} & \small{アリガトウ (\textit{arigatou}, thank you), ぃぃよ (\textit{iiyo}, no problem)} \\ \bottomrule
\end{tabular}
}
\vskip -1mm
\caption{List of annotation labels and examples for each phenomenon.}
\vskip -1mm
\label{tab:label_example}
\end{table*}

\paragraph{(ii) Extraction and normalization of targeted expressions}

We used crowdsourcing to add annotations to the MTNT dataset.
Considering the difficulty and inter-annotator agreement of the task, we divided the whole process into three subtasks: (i) annotating phenomena labels, (ii) extracting targeted expressions, and (iii) normalizing the expressions.
To ensure the quality, we assigned five workers per sentence for all tasks and retained the result only if a majority of workers answered the same.

Firstly, we asked crowdworkers to classify the source (Japanese) sentences with the above definitions~(Figure~\ref{fig:dataset_creation} Step1).
A question consists of four yes-no questions, each of which corresponds to one of the four phenomena.
We asked if there exist one or more expressions presenting each phenomenon for each sentence in the dataset.\footnote{Note that a sentence could be given more than one label. These sentences are treated differently according to the label which we focus on.}

Then, we associated the labels with corresponding expressions in a sentence.
More specifically, we designed a task to extract up to five expressions from a source sentence for each (source sentence, label) pair.
Also, we asked crowdworkers to extract translation of the targeted expressions, i.e.,\ the alignment from the target language~(Figure~\ref{fig:dataset_creation} Step2).
To avoid some sentences from being overrated, we discarded sentences having more than one expression with the same label.

Finally, to create contrastive input from raw noisy sentences, we asked crowdworkers to normalize the extracted expressions in the source language~(Figure~\ref{fig:dataset_creation} Step3).
The process of normalization stands for rewriting an expression to its canonical form in the dictionary, namely applying an inverse transformation to remove the reason for the classification.
For instance, the workers are to normalize an expression アプデ (\textit{apude}, an example of \textit{Abbreviated Noun} in Table~\ref{tab:label_example}) to アップデート (\textit{update}) by resolving abbreviation.
Another example from \textit{Colloquial Expression} is to normalize ねむーーい (\textit{nem\=ui}, sleeeepy) to ねむい (\textit{nemui}, sleepy) by removing unnecessary prolonged sound.
Here, the word is more commonly written in \textit{kanji} characters as 眠い (\textit{nemui}, sleepy) than in \textit{hiragana} characters (as in the example), however, the workers are instructed not to normalize the word in two stages because it is outside the scope of \textit{Colloquial Expression}.
On the other hand, if the given expression, ねむい (\textit{nemui}, sleepy) was originally in the text, it is counted as a \textit{Variant} and will be normalized to its \textit{kanji} notation.
We skipped this step for \textit{Proper Noun} because there is no concept of \textit{canonical form} for proper nouns.

We created our phenomenon-wise dataset in the form of quadruple consists of (original source sentence, normalized source sentence, alignment, target sentence) by replacing the extracted expressions with their normalized counterparts.
Table~\ref{tab:dataset_examples} shows some examples from our final dataset.
Also, we provide basic statistics of the dataset in Table~\ref{tab:stat_pheno_wise}.

\section{Translation Models}

\begin{table}[t]
\begin{tabular}{cc}
\begin{minipage}[t]{0.4\hsize}
\scalebox{0.82}{
\begin{tabular}{@{}ll@{}}
\toprule
\rownumber & \textit{Abbreviated Noun}  \\ \midrule
Orig. (Ja) & 地味な\textbf{アプデ}(\textit{apude}, update)だが \\
Norm. (Ja) & 地味な\textbf{アップデート}(\textit{update})だが \\
Ref. (En) & That's a plain \textbf{update} though \\ \bottomrule
\end{tabular}
}
\end{minipage} &
\begin{minipage}[t]{0.4\hsize}
\scalebox{0.82}{
\begin{tabular}{@{}ll@{}}
\toprule
\rownumber & \textit{Colloquial Expression} \\ \midrule
Orig. (Ja) & ここまで描いて飽きた、\textbf{かなちい} (\textit{kanachii}, sad) \\
Norm. (Ja) & ここまで描いて飽きた、\textbf{かなしい} (\textit{kanashii}) \\
Ref. (En) & Drawing this much then getting bored, how \textbf{sad}. \\ \bottomrule
\end{tabular}
}
\end{minipage} 
\end{tabular}
\vskip -1mm
\caption{Examples of original sentence (Orig.), normalized sentence (Norm.), and reference sentence (Ref.) in our dataset.}
\label{tab:dataset_examples}
\end{table}

\begin{table}[t]
\centering
\scalebox{0.9}{
\begin{tabular}{lrrr}
\toprule
Dataset & \# sent. & \# unique expressions (ratio) & average edit distance \\ \midrule
\textit{Proper Noun} & 943 & 747 (79.2\%) & - \\
\textit{Abbreviated Noun} & 348 & 234 (67.2\%) & 5.04 \\
\textit{Colloquial Expression} & 172 & 153 (89.0\%) & 1.77 \\
\textit{Variant} & 103 & 97 (94.2\%) & 3.42 \\ \bottomrule
\end{tabular}
}
\vskip -1mm
\caption{Basic statistics of our phenomenon-wise dataset.}
\vskip -3mm
\label{tab:stat_pheno_wise}
\end{table}

We prepared five in-house models with different size of training data and preprocessing methods for our experiments. %
The smaller model (\textsc{Small}) was trained on the data offered in the WMT 2019 shared task for machine translation robustness, namely TED talks, KFTT (Kyoto Free Translation Task), and JESC (Japanese-English Subtitle Corpus).
The MTNT dataset was also available in the task, but we didn't include any of the sentence pairs to train our models.
We replaced \textit{emojis} and emoticons with placeholders following a previous study by Murakami et al.~\shortcite{murakami:2019:wmt}.
In addition, we replaced possible usernames with regular expressions.
We offered this model to see whether or not the phenomena would become less problematic with increasing amount of training data.

For the other four models, we additionally used JParacrawl v2.0~\cite{morishita:2020:lrec}, one of the largest parallel corpora available in Japanese-English.
The larger model (\textsc{Large}), is only different in the size of training data from the \textsc{Small}.
We applied Byte-Pair-Encoding (BPE) models~\cite{sennrich:2016:acl} with a joint vocabulary of 32,000 for these models using the \texttt{sentencepiece} toolkit~\cite{kudo:2018:emnlp}.

The character-based model (\textsc{Char}) is different from the two models in the way of segmentation.
The model translates a sequence of characters in the source language into another sequence of characters in the target language~\cite{wang:2015}.
Durrani et al.~\shortcite{durrani:2019:naacl} pointed out that character-based models are more robust to noisy text than BPE-based models.
We revisit the issue of segmentation to see if the model is also good at handling UGC.
We shared the vocabulary between two languages in this setting as well to expect the model to capture copying behavior.

For the pronunciation-based model (\textsc{Pron}), we applied a unique preprocessing method to the source (Japanese) sentence.
More specifically, we first applied the \texttt{MeCab} toolkit~\cite{kudo:2004:emnlp}, a Japanese morphological analyzer, with naist-jdic for the dictionary to obtain the pronunciation of each morpheme in the sentences.
We can transliterate any words in Japanese by using phonetic symbols such as \textit{hiragana} and \textit{katakana} characters.
Since the \texttt{MeCab} toolkit outputs the pronunciation in \textit{katakana} characters by default, we simply concatenated them to create a fully pronunciation-based corpus.
We prepared this model with the expectation to improve the robustness against \textit{Variant} expressions.
More specifically, we aimed at absorbing the orthographic variations caused by \textit{hiragana}-\textit{katakana} confusion, which is a part of \textit{Variant}.
Also, previous study suggests that phonetic information is highly useful to resolve homophone noise~\cite{liu:2019:acl}.

Finally, we prepared the concatenated model (\textsc{Cat}), trained on a joined corpus for the \textsc{Large} and the \textsc{Pron}.\footnote{We also tried combining two sentences with delimiter tokens \texttt{<sep>} like the \textit{paste} command in the Unix-like operating systems, but we could not see any meaningful results from the model.}
In this setting, we converted the transliterated part into \textit{hiragana} characters and applied the same BPE model as used in the \textsc{Large} to the whole corpus.
We expect the model to learn robust representations by forcing it to produce the same translation from the original source sentence and its transliterated counterpart.
We used transformer-base architecture~\cite{vaswani:2017:nips} implemented in the \texttt{fairseq} toolkit~\cite{ott:2019:naacl} and hyperparameters proposed by Murakami et al.~\shortcite{murakami:2019:wmt} for all models.
The size of the training data was 3.9\,M for the \textsc{Small}, 14.0\,M for the \textsc{Large}, \textsc{Char} and \textsc{Pron}, and 28.0\,M for the \textsc{Cat}.

In addition to the in-house models, we also investigated the impact of the phenomena on two widely used MT systems, namely, Google Translate\footnote{\url{https://translate.google.co.jp}} and DeepL Translator. \footnote{\url{https://www.deepl.com/translator}}\footnote{The results are as of June 10, 2020.}
These systems are expected to be more robust against UGC because they are by nature exposed to user input.
By conducting experiments on such systems, we reveal the presence of phenomena with impending needs for improvement, and also confirm the usefulness of normalization as one of the tricks users can do.

\section{Phenomenon-wise Evaluation}
We provide an overview of the current state of NMT by evaluating the performance of both in-house models and off-the-shelf systems on the proposed phenomenon-wise dataset.
We fed both the original and normalized sentences to the models and measured the difference of single reference BLEU between them.
Since the only difference between the two sentences is the presence of the corresponding phenomenon, our dataset ensures that a phenomenon degrades the models more significantly if there is a larger gain of BLEU score after replacement.
We also calculated the ratio of correctly translated expressions, i.e., the accuracy, before and after normalization.
While the BLEU-based method enables us to measure the relative change in fluency from the sentence level, the accuracy-based method is rather aimed at evaluating the models locally.
We used these two measures supplementarily to investigate more closely what becomes an obstacle to current MT systems.

\subsection{Quantitative analysis}\label{sec:result_quantitative}

\begin{table}[t]
\centering
\resizebox{\textwidth}{!}{%
\begin{tabular}{@{}lrrrrrrrrrr@{}}
\toprule
 & \multicolumn{2}{c}{\textsc{Small}} & \multicolumn{2}{c}{\textsc{Large}} & \multicolumn{2}{c}{\textsc{Char}} & \multicolumn{2}{c}{\textsc{Pron}} & \multicolumn{2}{c}{\textsc{Cat}} \\ \cmidrule(l){2-3} \cmidrule(l){4-5} \cmidrule(l){6-7} \cmidrule(l){8-9} \cmidrule(l){10-11} 
 & Orig. / Norm. & Diff. & Orig. / Norm. & Diff. & Orig. / Norm. & Diff. & Orig. / Norm. & Diff. & Orig. / Norm. & Diff. \\ \midrule
\textit{Abbreviated Noun} & 10.4 / 10.8 & \fc+0.4 & 14.5 / 14.4 & \fc-0.1 & 11.8 / 12.0 & \fc+0.2 & 10.2 / 10.9 & \fc+0.7 & 13.8 / 13.6 & \fc-0.2 \\
\textit{Colloquial Expression} & 11.9 / 12.7 & \fc+0.8 & 13.8 / 14.9 & \fc\textbf{+1.1} & 12.3 / 11.7 & \fc-0.6 & 10.4 / 11.5 & \fc\textbf{+1.1} & 13.9 / 14.7 & \fc+0.8 \\
\textit{Variant} & 10.4 / 10.9 & \fc+0.5 & 13.7 / 15.3 & \fc\textbf{+1.6} & 13.2 / 16.0 & \fc\textbf{+2.8} & 11.1 / 11.9 & \fc+0.8 & 13.3 / 15.7 & \fc\textbf{+2.4} \\ \bottomrule
\end{tabular}%
}
\vskip -1mm
\caption{BLEU scores measured with our dataset (in-house models). \scriptsize{* Orig. for original, Norm. for normalized sentences.}}
\vskip -1mm
\label{tab:bleu_in-house_ext}
\end{table}

\begin{table}[t]
\centering
\resizebox{\textwidth}{!}{%
\begin{tabular}{@{}lrrrrrrrrrr@{}}
\toprule
 & \multicolumn{2}{c}{\textsc{Small}} & \multicolumn{2}{c}{\textsc{Large}} & \multicolumn{2}{c}{\textsc{Char}} & \multicolumn{2}{c}{\textsc{Pron}} & \multicolumn{2}{c}{\textsc{Cat}} \\ \cmidrule(l){2-3} \cmidrule(l){4-5} \cmidrule(l){6-7} \cmidrule(l){8-9} \cmidrule(l){10-11}
 & Orig. / Norm. & Diff. & Orig. / Norm. & Diff. & Orig. / Norm. & Diff. & Orig. / Norm. & Diff. & Orig. / Norm. & Diff. \\ \midrule
\textit{Proper Noun} & 34.3 / - & - & 49.7 / - & - & 47.1 / - & - & 43.2 / - & - & 48.0 / - & - \\
\textit{Abbreviated Noun} & 24.1 / 30.5 & \fc+6.4 & 33.6 / 33.0 & \fc-0.6 & 34.2 / 34.8 & \fc+0.6 & 30.2 / 31.3 & \fc+1.1 & 34.2 / 33.0 & \fc-1.2 \\
\textit{Colloquial Expression} & 18.0 / 23.8 & \fc+5.8 & 14.5 / 24.4 & \fc+9.9 & 17.4 / 21.5 & \fc+4.1 & 8.7 / 30.2 & \fc\textbf{+21.5} & 15.7 / 32.6 & \fc\textbf{+16.9} \\
\textit{Variant} & 15.5 / 35.0 & \fc\textbf{+19.5} & 13.6 / 38.8 & \fc\textbf{+25.2} & 13.6 / 34.0 & \fc\textbf{+20.4} & \textbf{25.2} / 35.9 & \fc\textbf{+10.7} & \textbf{26.2} / 35.0 & \fc+8.8 \\ \bottomrule
\end{tabular}%
}
\vskip -1mm
\caption{Accuracy (\%) measured with our dataset (in-house models). \scriptsize{* Orig. for original, Norm. for normalized sentences.}}
\vskip -3mm
\label{tab:acc_in-house_ext}
\end{table}

\paragraph{In-house models}
Table~\ref{tab:bleu_in-house_ext} and~\ref{tab:acc_in-house_ext} show the BLEU scores and the accuracy, respectively, for the in-house models.
The results showed that the scores were constantly improved after normalization for the \textsc{Small}, which indicates that all of the targeted phenomena may adversely affect the model to some extent.
However, there seems to be a clear difference in the trend between \textit{Proper Noun}, \textit{Abbreviated Noun} and the other two UGC-specific phenomena.
First, the accuracy with original sentences for the \textit{Proper Noun} and \textit{Abbreviated Noun} increased with the size of training data, while we observed a slight drop for the other two.
Also, the gain from normalization for the \textit{Abbreviated Noun} was exceptionally high in the \textsc{Small}.
It is also notable that the difference scores for the \textit{Colloquial Expression} and \textit{Variant} were even larger in the \textsc{Large} than in the \textsc{Small}.
These figures support that we need special treatment beyond collecting massive training data to further improve MT systems on UGC. 

From the point of tokenization, the \textsc{Char} could not outperform the \textsc{Large} in all phenomena with the BLEU scores.
However, we could see an improvement of 5.8 points in the difference of accuracy for the \textit{Colloquial Expression}, showing its high robustness against the phenomenon.
We speculate that this might result from the small edit distance in the \textit{Colloquial Expression} dataset.
Similar to typographical errors in alphabetical languages, character-based models seem to prove their true worth with phenomena for which surrounding characters become an important clue.
On the other hand, it was surprising that \textit{Variant}, which is an instance of orthographic variations, was not treated well by the model.
This might result from the fact that \textit{Variant} is rather a word-level phenomenon unlike typographical errors, which are in most cases limited within several characters.

The \textsc{Pron} also performed poorly with the BLEU scores.
However, it is notable that the model showed the smallest difference score for the \textit{Variant} among four models trained on the larger data.
Here, we refer to Table~\ref{tab:acc_in-house_ext} for the translation accuracy of the models.
While the accuracy for the \textit{Variant} after normalization stayed almost the same for all five models, the accuracy for the original sentences attained by the \textsc{Pron} went more than 10 points higher than that by the \textsc{Large} and the \textsc{Char}.
The results indicate that the decrease in difference is not brought by the limited expressiveness of phonetic symbols but by the increasing capacity to handle non-standard input.
We might discard the model for its low BLEU scores without our dataset, but our phenomenon-wise dataset provides a new axis to the evaluation, discovering the possibility of the model.

The \textsc{Cat} seems to be a better alternative to the \textsc{Pron}.
The model showed a relatively small drop in the BLEU scores from the \textsc{Large} (Table~\ref{tab:bleu_in-house_ext}), and also benefited from the robust representations of the pronunciation-based corpus.
The model reached 26.2\% accuracy for the \textit{Variant}, which is significantly higher than 13.6\% by the \textsc{Large} (Table~\ref{tab:acc_in-house_ext}).
Also, the accuracy for the \textit{Colloquial Expression} showed 8.2 points gain after normalization as compared to the \textsc{Large}.
This implies that the model could be more adaptive to the phenomenon with proper preprocessing.
We speculate that one reason for the improvement comes from the increasing capacity of the \textsc{Cat} to treat unexpected segmentation caused by \textit{hiragana} characters.
In Japanese, most of the highly-frequent function words consist of a few \textit{hiragana} characters.
Sasano et al.~\shortcite{sasano:2013:ijcnlp} pointed out that expressions in \textit{hiragana} characters are more likely to combine each other to produce these function words than kept as single words.
The idea of mixing a pronunciation-based corpus forces a model to produce correct output from unexpectedly segmented, difficult sequences.
The results suggest the importance of deep consideration for possible perturbations from the viewpoint of linguistic phenomena to better handle UGC.

Overall, \textit{Proper Noun} was handled relatively well by all in-house models as compared with the other three phenomena.
The results also showed that BPE-based models (\textsc{Large} and \textsc{Pron}) performed slightly better with proper nouns than character-based models. 
On the other hand, the scores for the \textit{Abbreviated Noun} were rather inconsistent: the differences even went into minus in some models.
However, the result does not necessarily mean that the phenomenon is less important to cope with.
To investigate the effect of \textit{Abbreviated Noun} more deeply, we conducted an additional experiment to subdivide the dataset into several groups.{\footnote{See Appendix~\ref{appendix:subdivision_exp} for the detail of the experiment.}}
The result showed that there were roughly two types of expressions for the phenomenon, namely the alphabetical acronyms and the others, and the behavior of the models was completely different from each other.
The process of normalization unnecessarily led a model to explain the acronyms redundantly to induce a drop in accuracy.
It might be better to exclude these expressions from our \textit{Abbreviated Noun} dataset for more precise evaluation.

\begin{table}[t]
\centering
\resizebox{\textwidth}{!}{%
\begin{tabular}{@{}lrrrrrrrr@{}}
\toprule
 & \multicolumn{4}{c}{BLEU} & \multicolumn{4}{c}{Accuracy (\%)} \\ \cmidrule(l){2-5} \cmidrule(l){6-9}
 & \multicolumn{2}{c}{Google Translate} & \multicolumn{2}{c}{DeepL Translator} & \multicolumn{2}{c}{Google Translate} & \multicolumn{2}{c}{DeepL Translator} \\ \cmidrule(l){2-3} \cmidrule(l){4-5} \cmidrule(l){6-7} \cmidrule(l){8-9}
 & Orig. / Norm. & Diff. & Orig. / Norm. & Diff. & Orig. / Norm. & Diff. & Orig. / Norm. & Diff. \\ \midrule
\textit{Proper Noun} & - / - & - & - / - & - & 55.2 / - & - & 50.5 / - & -  \\
\textit{Abbreviated Noun} & 14.6 / 15.0 & \fc+0.4 & 16.3 / 16.2 & \fc-0.1 & 41.1 / 36.8 & \fc-4.3 & 39.1 / 37.9 & \fc-1.2 \\
\textit{Colloquial Expression} & 14.4 / 16.0 & \fc\textbf{+1.6} & 15.6 / 15.8 & \fc+0.2 & 19.2 / 26.2 & \fc+7.0 & 22.7 / 28.5 & \fc+5.8 \\
\textit{Variant} & 15.3 / 17.6 & \fc\textbf{+2.3} & 14.4 / 15.2 & \fc+0.8 & 23.3 / 37.9 & \fc\textbf{+14.6} & 18.4 / 35.0 & \fc\textbf{+16.6} \\ \bottomrule
\end{tabular}%
}
\vskip -1mm
\caption{BLEU scores and accuracy (\%) measured with our dataset (off-the-shelf systems). \scriptsize{* Orig. for original, Norm. for normalized sentences.}}
\vskip -3mm
\label{tab:res_off-the-shelf}
\end{table}

\paragraph{Off-the-shelf systems}
It is worth surprising that even the off-the-shelf systems performed poorly with our \textit{Variant} dataset (Table~\ref{tab:res_off-the-shelf}).
The systems dropped more than 10 points in accuracy when faced with the original sentences, and showed large differences in the BLEU scores as well.
Also, the result is quite suggestive in that a system better at BLEU scores is not always better at handling specific phenomena.
For instance, the accuracy for the \textit{Abbreviated Noun} dataset with DeepL Translator was 2 points lower than Google Translate, but the system showed 1.7 points higher BLEU score.
We speculate that this might have been caused by the different behaviors of the two systems.
In our experiments, DeepL Translator tended to ignore uncommon phrases to keep the overall translation fluent, but Google Translate endeavored to provide some output even if phrases were confusing to the model.
Practically, the preference over high-precision systems or high-recall systems depends on the application for which the translation is used.
The two-way evaluation, from the BLEU scores and the accuracy, could be of great help for us to make a decision about what models to deploy.

\subsection{Qualitative analysis}\label{sec:result_qualitative}

\begin{table}[t]
\centering
\scalebox{0.9}{
\begin{tabular}{@{}ll@{}}
\toprule
(a) \textit{Variant} &  \\ \midrule
Source & \{ぎゃくたい / 虐待 (\textit{gyakutai}, abuse)\} だ \\
\textsc{Large}$_{orig}$ & I want to do it! \\
\textsc{Cat}$_{orig}$ &  It's \textbf{abuse}! \\
\textsc{Large}$_{norm}$ & It's \textbf{abuse}! \\ 
Reference & It's abuse! \\ \midrule
(b) \textit{Abbreviated Noun} &  \\ \midrule
Source & 進化する \{サバゲー (\textit{sabag\=e}, survival game) / サバイバルゲーム (\textit{survival game})\} \\
\textsc{Large}$_{orig}$ & The evolving mackerel game. \\
\textsc{Cat}$_{orig}$ & Evolving sabage. \\
\textsc{Large}$_{norm}$ & Evolving \textbf{Survival Game} \\
Reference & An evolving survival game. \\ \midrule
(c) \textit{Proper Noun} &  \\ \midrule
Source & 平昌 (\textit{Pyongchang}) で「米日VS南北」の戦いが始まる \\
\textsc{Small} & In the Heisho era, the battle of 'South and South America' began. \\
\textsc{Large} & The Battle of 'America-Japan VS North-South' begins in \textbf{Pyeongchang} \\
Reference & The "US and Japan vs. North and South Korea" battle has begun in Pyeongchang. \\ \bottomrule
\end{tabular}%
}
\vskip -1mm
\caption{Output examples from our in-house models. \scriptsize{*\{original expression / normalized expression\}}}
\vskip -3mm
\label{tab:output_examples}
\end{table}

We also analyzed qualitatively how translations generated by the models were changed after normalization.
Table~\ref{tab:output_examples} shows some examples of the output from our in-house models.

Example (a) is a case where the output was improved by replacing the \textit{hiragana} expression ぎゃくたい (\textit{gyakutai}, abuse) with its common notation in \textit{kanji}, 虐待 (\textit{gyakutai}).
In this case, the \textsc{Large} mistakenly output the phrase \textit{want to} when we fed the original source sentence.
This might have resulted from the fact that the original expression was overly segmented into four parts with our BPE model.
Here, the presence of a segmented prefix たい (\textit{tai}), a highly-frequent auxiliary verb often combined with other verbs to show one's desire, possibly worked badly to produce the wrong output.
On the other hand, though the input was the same, the \textsc{Cat} could produce a correct translation, \textit{abuse} for the original expression.
In most case, the character preceding the auxiliary verb たい (\textit{tai}) generates \textit{i} or \textit{e} sound, such as したい (\textit{shitai}, want to do) and 食べたい (\textit{tabetai}, want to eat).
The pronunciation-based corpus might have provided enough false examples to learn this rule, resulted in the improvement.

Though \textit{Variant} is one of the phenomena specific to languages with various writing systems, similar problems are actually observed in other languages as well.
For example, the negative effects caused by some types of typographical errors can be explained in the same way as the example above.
It is a challenge how we obtain correct translation in case that an erroneous expression is partially associated with other words.

The next example (b) is from our \textit{Abbreviated Noun} dataset.
In this example, the \textsc{Large} could not produce the correct translation for the original expression サバゲー (\textit{sabag\=e}, survival game), and mistakenly treated the word as a combination of the two words, サバ (\textit{saba}, mackerel) and ゲー (\textit{g\=e}, game).
The \textsc{Cat} also suffered from translating the expression, but it instead transliterated the word into the alphabet.
The result implies that the model captures character-level cooccurrence inside a word better than naive models: mackerels usually do not appear in a game.
Also, we found an interesting example where an abbreviated word could be interpreted differently according to the context (生保, \textit{seiho}, life insurance or life security).
It is important to capture the context not only inside but outside a word to further improve the models.

Finally, in the example (c), the expression 平昌 (\textit{Pyongchang}) was correctly handled by the \textsc{Large}, while the \textsc{Small} could not.
Though it is not unnatural to conclude that the increasing capacity of treating \textit{Proper Noun} resulted from the large corpus on which the model was trained, we believe it is not a sufficient condition to explain the consequence.
An observation behind is that the term \textit{Pyongchang} became popular after the Olympics was held there in 2018.
The corpora we used for training the \textsc{Small} were no newer than 2018, and that possibly resulted in fewer occurrences of the term.
To create truly robust systems against \textit{Proper Noun}, we believe it is necessary to divide corpora chronologically to measure the generalization ability against nouns that appear only in the test data.
However, we believe our dataset could be of some help to evaluate models' performance against the phenomenon, considering the fact that it is quite unrealistic to keep a test data always newer than any training data.

\section{Correlation with Human Evaluation}\label{sec:discussion}

To demonstrate a potential use case of our phenomenon-wise dataset, we conducted an additional experiment, where we reassessed the systems submitted to the WMT 2019 robustness shared task in a phenomenon-wise manner.
We downloaded five official submissions for the blind test\footnote{The data used for ranking systems in the shared task. It was kept blind to participants until the evaluation period ends.} portion of the MTNT dataset.\footnote{\url{http://matrix.statmt.org/matrix/systems_list/1917}}
We then extracted the intersection between the blind test data and our phenomenon-wise dataset, obtaining 136 sentences for \textit{Proper Noun}, 48 for \textit{Abbreviated Noun}, 21 for \textit{Colloquial Expression}, and 11 for \textit{Variant}.
The task organizer also provides the results of human judgment of all submissions for each sentence~\cite{li:2019:wmt}, where three human raters were instructed to rate each translation on a scale from 1 (completely incorrect) to 100 (accurate).
For each of the five submissions, we averaged all the human ratings for each sentence in the phenomenon subset and investigated the correlation between the averaged human ratings and the phenomenon-wise accuracy.

\begin{figure}[t]
\begin{tabular}{cccc}
\begin{minipage}[t]{0.22\hsize}
\centering
\includegraphics[keepaspectratio, scale=0.295]{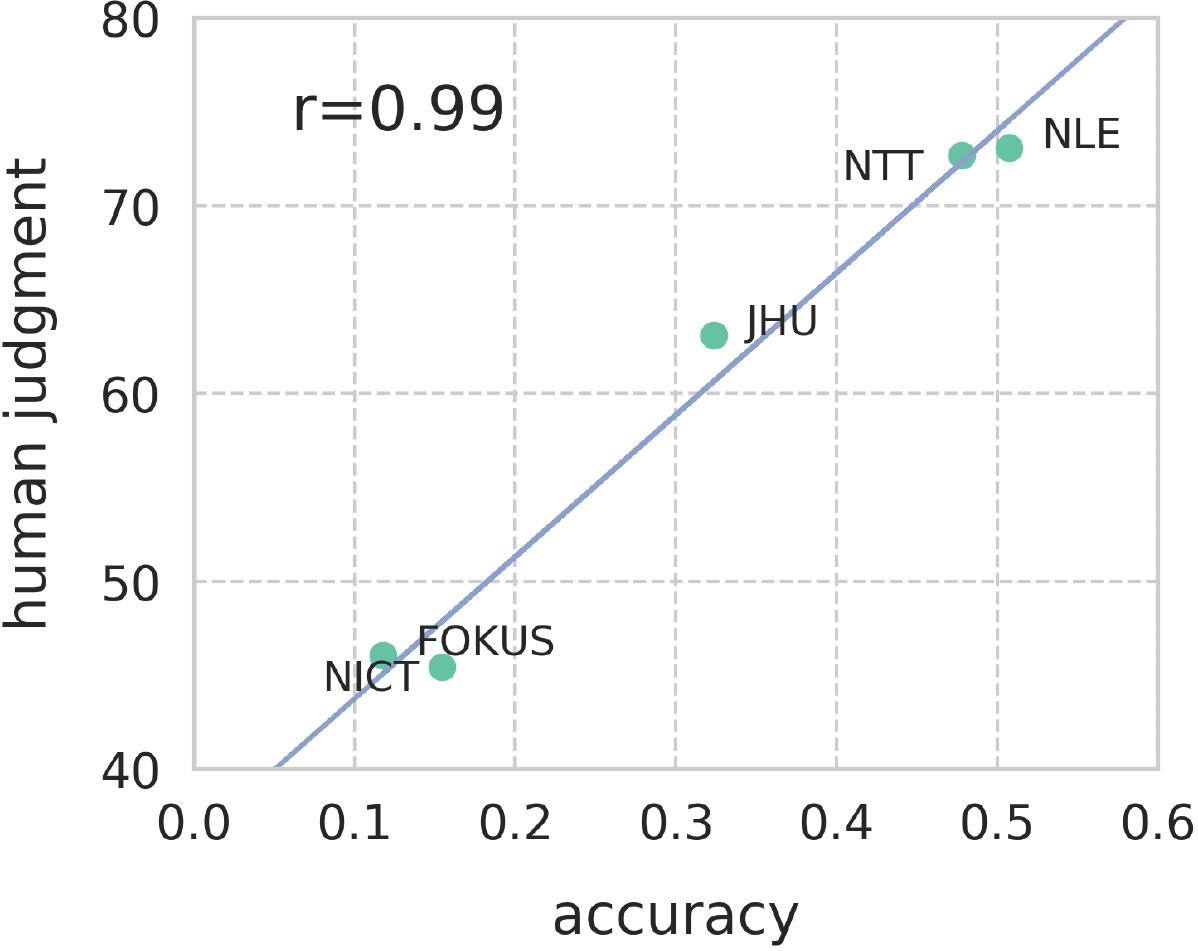}

\small{(a) \textit{Proper Noun}}
\end{minipage} &
\begin{minipage}[t]{0.22\hsize}
\centering
\includegraphics[keepaspectratio, scale=0.295]{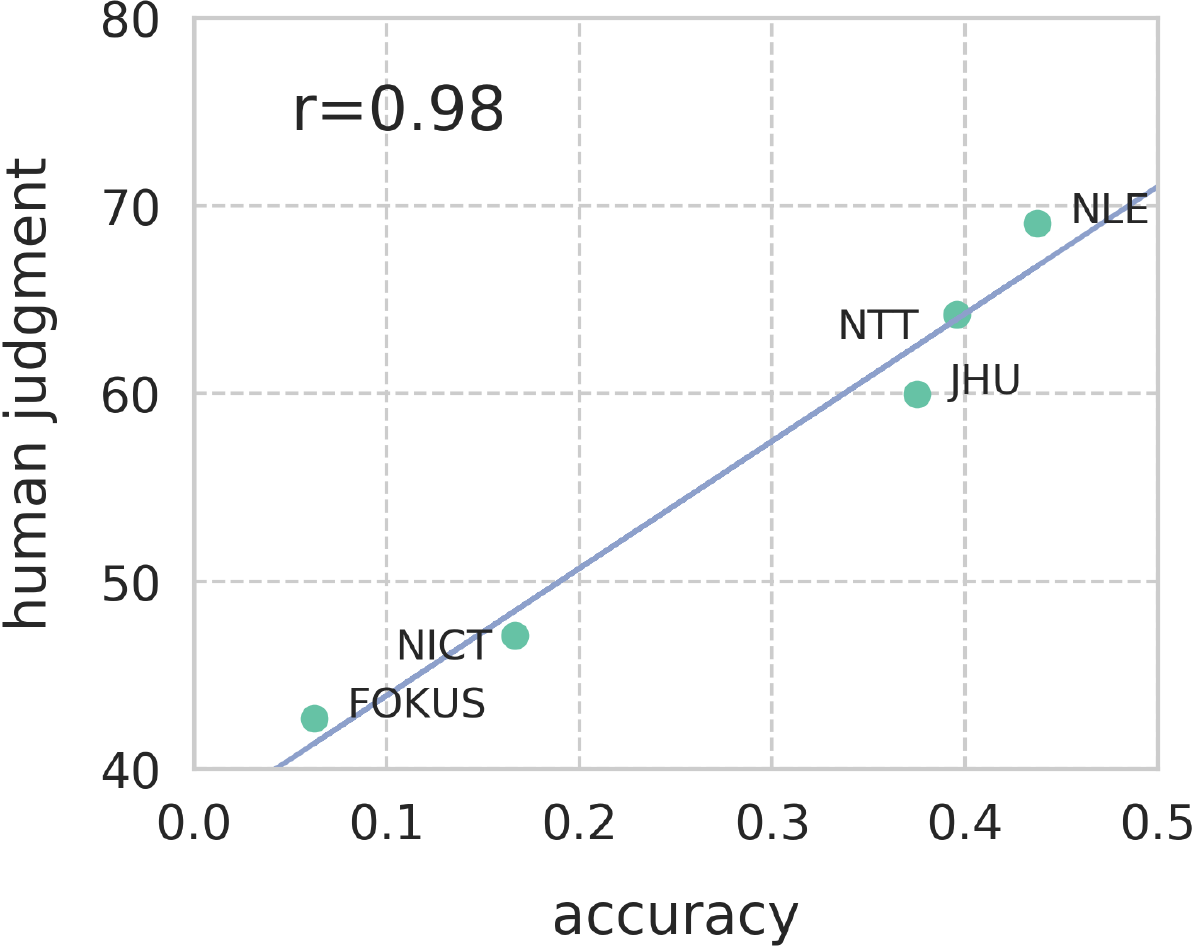}

\small{(b) \textit{Abbreviated Noun}}
\end{minipage} &
\begin{minipage}[t]{0.22\hsize}
\centering
\includegraphics[keepaspectratio, scale=0.295]{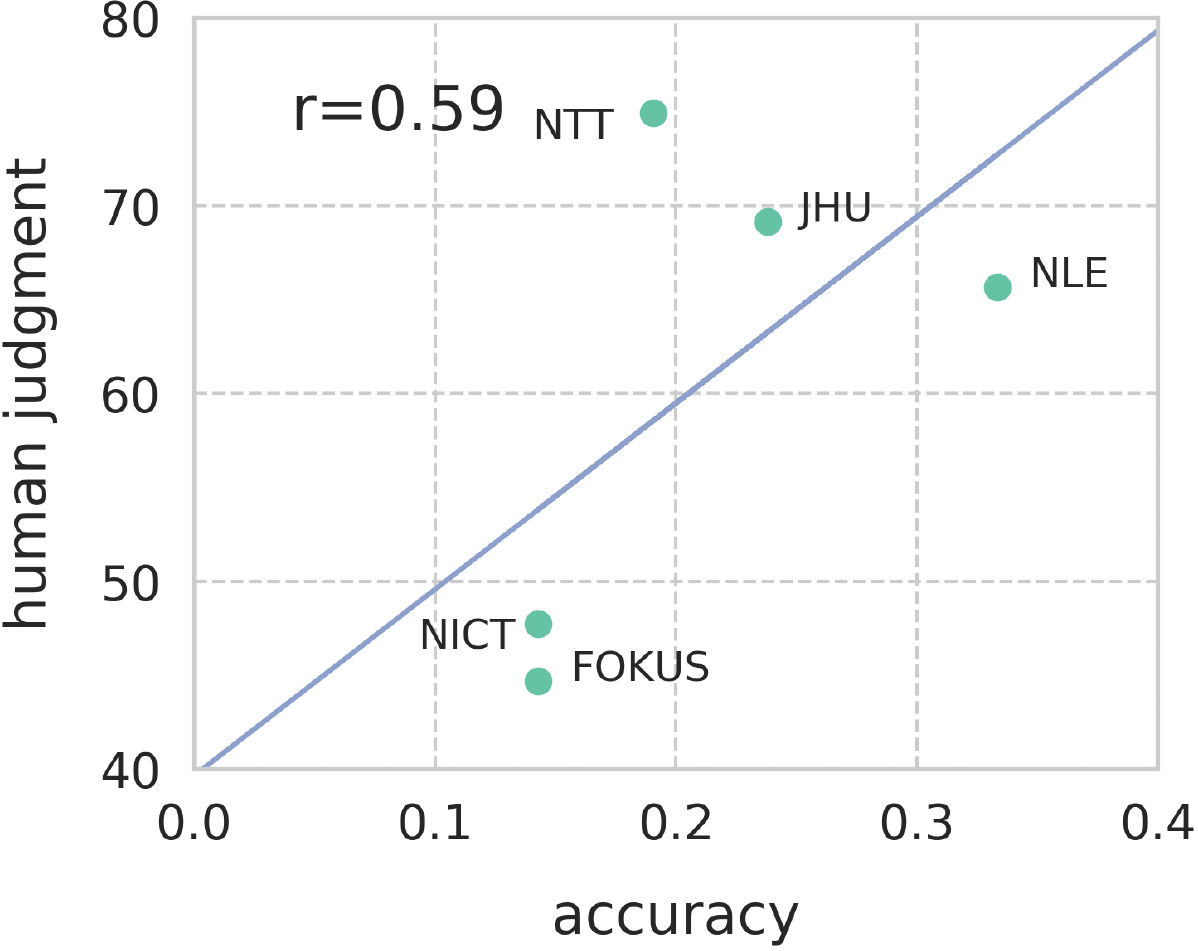}

\small{(c) \textit{Colloquial Expression}}
\end{minipage} &
\begin{minipage}[t]{0.22\hsize}
\centering
\includegraphics[keepaspectratio, scale=0.295]{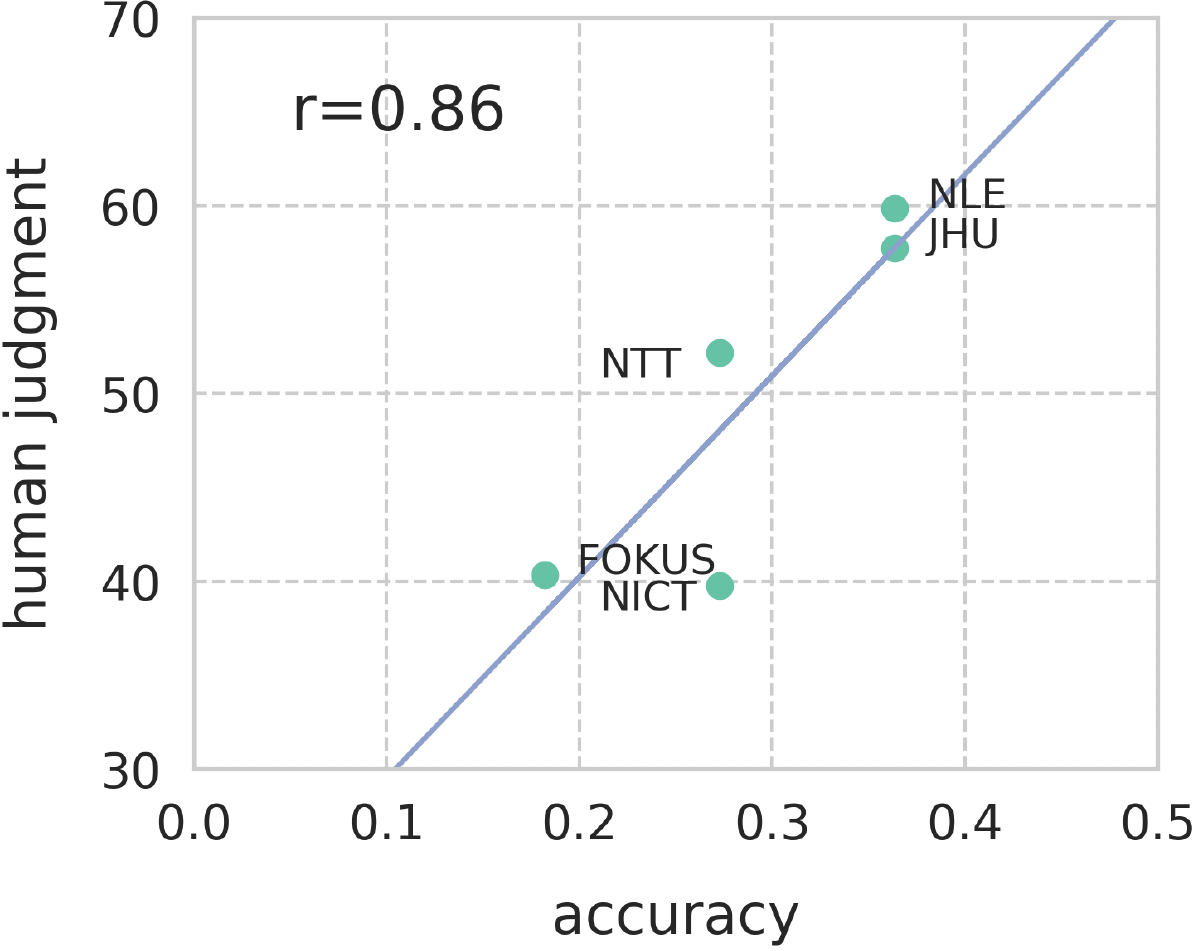}

\small{(d) \textit{Variant}}
\end{minipage}
\end{tabular}
\caption{Correlation between the accuracy and human judgment scores for each phenomenon (WMT submitted systems). The $r$-value is Pearson's correlation coefficient.}
\vskip -2mm
\label{fig:corr}
\end{figure}

From the results in Figure~\ref{fig:corr}, we could see that the accuracy for our \textit{Proper Noun} and \textit{Abbreviated Noun} dataset strongly correlated to the human judgment scores with $r > 0.9$.
This is worth surprising because we have no access to the whole sentences but only to the targeted expressions in our accuracy-based method.
The result suggests that the two phenomena are key factors for humans in evaluating overall translation quality.
One reason might be that humans can easily tell whether the translated sentences include these nouns or not.
This implies that undertranslation of words for these two phenomena could bring a more serious impact on human judgment.
We believe that the accuracy can be used as a strong signal for estimating human judgment scores, when combined with traditional evaluation metrics such as BLEU.

\section{Conclusion}
We proposed a novel dataset designed for phenomenon-wise evaluation in Japanese-English translation.
In this research, we focused on four linguistic phenomena commonly seen on User-Generated Contents, namely \textit{Proper Noun}, \textit{Abbreviated Noun}, \textit{Colloquial Expression}, and \textit{Variant}.

Using our dataset, we analyzed how current MT systems are negatively affected by the presence of the phenomena.
The result showed that \textit{Variant} is one of the phenomena that significantly degrade the model's performance including widely used, strong off-the-shelf systems.
This implies that collecting massive training corpora is not a sufficient condition to handle these peculiar inputs, and we need special treatment against them to further improve MT systems.

We also analyzed the correlation between human judgment and translation accuracy scores from our dataset by using official submissions from the WMT 2019 shared task.
From the experiments, we confirmed that the accuracy-based scores from our dataset strongly correlated with human judgment, showing its potential to reduce the cost of the evaluation.

We made our dataset publicly available for further development in MT systems\footnote{\url{https://github.com/cl-tohoku/PheMT}}.
As future work, we would like to consider new model architectures or data preprocessing methods to improve performance against specific phenomena using our dataset.

\newpage
\bibliographystyle{coling}
\bibliography{coling2020}

\newpage
\appendix

\begin{table*}[t]
\centering
\scalebox{0.85}{
\begin{tabular}{cl}
\toprule
\bf Score & \bf Definition \\ \midrule
5  & translations that conveys the meaning completely and fluent as target language sentence \\
\rowcolor{gray!7}
4  & translations that does not show any lack of information, but highly Translationese (verbatim) \\ \midrule
3 & translations that has locally untranslated / mistranslated parts, but acceptable \\
\rowcolor{gray!7}
2 & translations that has phrase, sentence-level mistranslation, or based on different interpretation \\
1 & translations that is complete nonsense \\ \bottomrule
\end{tabular}%
}
\vskip -1mm
\caption{Criterion for appropriateness score annotation}
\vskip -2mm
\label{tab:appropriateness_criterion}
\end{table*}

\section{Preliminary experiment of appropriateness score annotation}
\label{appendix:appropriateness_exp}

To ensure the quality of the resulting dataset, we first applied some basic rule-based filtering to the corpus.
More specifically, we removed (i) sentences including inappropriate words using a predefined word list\footnote{\url{https://github.com/1never/open2ch-dialogue-corpus}. The list was created for dialog corpus filtering.}, (ii) pairs having identical source and target sentences, (iii) duplicates, and (iv) sentences consisting of 1 word, or more than 80 words.
Then, we designed a task to annotate the \textit{appropriateness} of the translation for each sentence.
The task was aimed to classify the source and target sentence pairs on the Likert scale having scores ranging from 1 (very poor) to 5 (excellent).
To define the criterion, we followed the common practice of assessing machine-translated output from two perspectives: adequacy and fluency~\cite{white:1994:amta}.
However, we added some modifications because the translations to be evaluated were human-generated.
The criterion for each grade is given in Table~\ref{tab:appropriateness_criterion}.

Since it requires a highly advanced understanding of the source language (Japanese) to correctly capture the meaning of sentences in UGC, we asked ten native speakers of Japanese with high English proficiency to annotate scores in this task.\footnote{We set the standard reward for each worker to 20,000 yen, approx.\ 185 dollars with the exchange rate as of June 2020. We selected workers who had rich experience in translation or had equivalent skills, from more than 80 applicants.}
We allocated three different workers per sentence, and averaged these scores to obtain the final score.
We filtered out sentences by the threshold of 4.0 and retained only one reference with the highest appropriateness score per source sentence to prevent negative effects caused by single reference BLEU: high precision for one reference may lower the precision for other references.

Figure~\ref{fig:appropriateness} shows the distribution of annotated appropriateness scores for each portion of the MTNT dataset.
There were 4152 sentences with an average score of 4.0 or more out of the 7273 annotated sentences.
The number of sentences discarded was large enough to support the necessity of pre-filtering by translation quality to assure the quality of our phenomenon-wise dataset.
The result also showed that the train and development portion of the dataset (blue and yellow bars in the figure) included more sentences in lower quality compared to the test and blind portion (green and red bars).
The difference was particularly clear in the range lower than the average score of 3.0 and higher than 4.0.
The number of sentences we kept for phenomena annotation was 3896, after retaining only one reference with the highest appropriateness score per source sentence.

\section{Subdivision of the Abbreviated Noun Dataset}
\label{appendix:subdivision_exp}

\begin{table}[t]
\centering
\resizebox{0.95\textwidth}{!}{%
\begin{tabular}{@{}cp{5em}p{9em}p{13em}rrr@{}}
\toprule
Group & Orig. & Norm. & Example & \# sents. & $\Delta$Acc. (\textsc{SMALL}) & $\Delta$Acc. (\textsc{LARGE}) \\ \midrule
1 & alphabetical & alphabetical & AI / artificial intelligence & 9 & -22.2 & -88.9 \\
\rowcolor{gray!7}
2 &  & \textit{katakana} & PC / パーソナルコンピューター (\textit{personal computer}) & 41 & -26.9 & -61.0 \\
3 &  & others & EU / 欧州連合 (\textit{oush\=urengou}, Europe Union) & 23 & -52.2 & -60.9 \\ \midrule
4 & mixed & first / last n characters & サンタ (\textit{Santa}) / サンタクロース (\textit{Santa Claus}) & 104 & +13.4 & +21.2 \\
\rowcolor{gray!7}
5 &  & combination of two first-n characters & アニオタ (\textit{aniota}, Anime nerds) / アニメオタク (\textit{animeotaku}) & 132 & +18.2 & +14.3 \\
6 &  & others & マック (\textit{makku}, McDonald's) / マクドナルド (\textit{makudonarudo}) & 39 & +23.1 & +10.2 \\ \midrule
 & overall &  &  & 348 & +6.4 & -0.6 \\ \bottomrule
\end{tabular}%
}
\vskip -1mm
\caption{Criterion and results of subdivided Abbreviated Noun dataset}
\vskip -2mm
\label{tab:abbrev_subdivision}
\end{table}
The results from Table~\ref{tab:acc_in-house_ext} and~\ref{tab:res_off-the-shelf} showed that \textit{Abbreviated Noun}, unlike other phenomena, did not affect the models in a negative way.
To further investigate the effect of the phenomenon, we additionally subdivided the dataset into six groups.
Table~\ref{tab:abbrev_subdivision} shows the criterion for each group and the difference in accuracy before and after normalization.
In the first three groups, wherein the original expressions were written in alphabetical acronyms, there was a severe drop of up to over 60\% accuracy with the \textsc{Large} after normalizing the expressions.
One reason to explain the result is that those acronyms are usually kept intact in the reference as they tend to be originally imported from the target language (English) to the source language (Japanese).
The process of normalization led models to unnecessarily explain terms redundantly, resulted in a drop in the accuracy, which is based on the exact match.
However, the output with normalized, expanded expression is not always a wrong translation.
For instance, we could see from the result that an expression \textit{DM} was translated as \textit{direct mail} after normalization.
It might be better to exclude these sentences from our \textit{Abbreviated Noun} dataset for precise evaluation.

On the other hand, expressions classified into the latter half of the groups seem to harm models significantly as normalization brought great improvement in the translation accuracy.
As we discussed in Section \ref{sec:result_qualitative}, this might result from increasing ambiguity caused by abbreviation.
We observed many expressions classified in these groups written in \textit{katakana} characters.
Among the four main types of characters used in Japanese, \textit{hiragana} and \textit{katakana} are less informative because of their characteristics as phonetic symbols.
The presence of abbreviations limits the number of accessible characters, and we believe it eventually imposes a deeper understanding of intra-sentential context on the models.

\end{document}